\UseRawInputEncoding
\documentclass[letterpaper, 10 pt, journal, twoside]{IEEEtran}

\usepackage{graphicx}
\usepackage{gensymb}
\usepackage{multirow}
\usepackage[font=small]{caption}
\usepackage{physics,amsmath}
\usepackage{amssymb}
\usepackage{tabularx}
\usepackage{booktabs}
\usepackage[breaklinks,colorlinks,citecolor=black]{hyperref}

\IEEEoverridecommandlockouts 

\title{
Using a Distance Sensor to Detect\\Deviations in a Planar Surface
}

\author{Carter Sifferman$^{1}$, William Sun$^{1}$, Mohit Gupta$^{1}$, and Michael Gleicher$^{1}$
\thanks{This paper was recommended for publication by Editor Tamim Asfour upon evaluation of the Associate Editor and Reviewers' comments.
This work was supported by Los Alamos National Lab and the Department of Energy, NSF CAREER Award 1943149, NSF grant CNS-2107060, and ONR grant N000142412155}
\thanks{$^{1}$The authors are with the Department of Computer Sciences,
         University of Wisconsin-Madison, Madison 53706, USA
        {\tt\footnotesize [sifferman|wsun|mohitg|gleicher]\@cs.wisc.edu}}%
}

\begin{document}
\maketitle
\begin{abstract}
We investigate methods for determining if a planar surface contains geometric deviations (\textit{e.g.}, protrusions, objects, divots, or cliffs) using only an instantaneous measurement from a miniature optical time-of-flight sensor. The key to our method is to utilize the entirety of information encoded in raw time-of-flight data captured by off-the-shelf distance sensors. We provide an analysis of the problem in which we identify the key ambiguity between geometry and surface photometrics. To overcome this challenging ambiguity, we fit a Gaussian mixture model to a small dataset of planar surface measurements. This model implicitly captures the expected geometry and distribution of photometrics of the planar surface and is used to identify measurements that are likely to contain deviations. We characterize our method on a variety of surfaces and planar deviations across a range of scenarios. We find that our method utilizing raw time-of-flight data outperforms baselines which use only derived distance estimates. We build an example application in which our method enables mobile robot obstacle and cliff avoidance over a wide field-of-view.

\vspace{1em}
\noindent
Project website: \url{https://cpsiff.github.io/using_a_distance_sensor/}
\end{abstract}


\vspace{-0.5em}
\section{Introduction}

\IEEEPARstart{O}{ptical} time-of-flight distance sensors are widely used in robotics to sense the distance to nearby objects for tasks such as obstacle avoidance~\cite{escobedo2021contact} or localization~\cite{karam2022microdrone, karimi2023slam}. These sensors are low-cost, low-power, and are available in low-resolution (\textit{e.g.}, 4x4 pixel) arrays requiring minimal data bandwidth. The distance estimates reported by these sensors are summaries over a wide (\textit{e.g.}, 10\textdegree) field-of-view per pixel, which is advantageous for some applications (\textit{e.g.}, conservative obstacle avoidance), but these distance estimates are ineffective at detecting small geometric deviations. In this work, we show that this limitation can be overcome by utilizing readily available raw time-of-flight information captured by these sensors. With this data, we are able to detect geometric deviations in a planar surface with more accuracy than is possible with distance estimates alone.

Our method utilizes raw time-of-flight data captured by consumer-grade \textit{time resolved} active time-of-flight sensors. These sensors operate by illuminating a wide (30\textdegree +) patch of the scene with a pulse of light, and capturing the intensity of light over time as it bounces back from the scene in a 1D temporal waveform called a \textit{transient histogram}~\cite{Callenberg2021CheapSPAD, jungerman2022}. These sensors are available for less than \$5 USD and are widely used in robotics applications~\cite{tsuji2022omnidirectional, niculescu2023nanoslam}. In addition to their low cost, they are very small ($<$20 mm$^3$) and low-power ($<$10 milliwatts per measurement)~\cite{VL53L8CH, TMF8820}. Typical applications do not utilize the transient histogram captured by these sensors, instead relying on a proprietary algorithm onboard the sensor to extract a single distance estimate per pixel. While this estimate is convenient for many tasks, it is not ideal for many others, as it obscures relevant information about the scene which is encoded in the shape and magnitude of the transient histogram.

\begin{figure}
    \centering
    \includegraphics{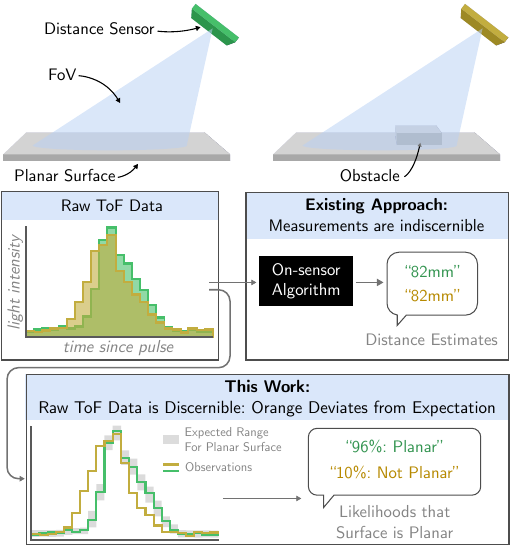}
    \caption{Our method uses raw time-of-flight data to detect deviations from planar surfaces (\textit{e.g.}, objects, divots, cliffs, or walls). It is able to do so more accurately than methods that utilize on-sensor distance estimates.}
    \vspace{-2em}
    \label{fig:teaser}
\end{figure}

In this work, we aim to detect geometric deviations on planar surfaces (\textit{e.g.}, objects, divots, cliffs, or walls). Our method assumes that the relative orientation and distance of the sensor to the planar surface remains fixed (\textit{i.e.}, only translation parallel to the surface is permitted), and our method requires a small dataset of measurements from the sensor to fit a surface model. We do not aim to detect the exact nature of the deviation; we only classify a measurement as ``planar'' or ``not planar'' (the latter class including planes viewed from a different orientation and/or distance). This capability is useful for robotics applications like mobile robot and drone navigation. It could also be useful for \textit{e.g.}, safely landing a drone on a flat and level surface or safely placing a cup of liquid on a clear and level portion of tabletop with a robot manipulator. Our method enables deviation detection in a very small size, weight, and compute footprint, making it particularly useful for resource-constrained scenarios like micro-drones, and for distributed sensing with sensors mounted at many points on a larger robot.

The key contributions of this work are: 1) an analysis of the problem of detecting deviations from a planar surface with a distance sensor, in which we identify the key ambiguity that makes the problem challenging, 2) a computationally lightweight method for detection of said deviations using raw sensor time-of-flight information, 3) characterization of the sensitivity and accuracy of said method compared to methods which use only sensor distance estimates, and 4) an example application in which our method enables mobile robot obstacle avoidance.

\section{Related Work}
\subsection{Obstacle Detection on Planar Surfaces}
Detecting obstacles on top of planar surfaces is a widely studied problem in computer vision and robotics \cite{badrloo2022image, yu2020study}. Approaches vary in their problem setup (per-frame detection or utilizing constrained robot motion) and imaging modality (RGB, depth from stereo, or structured light).

\subsubsection{Utilizing Robot Motion}
One class of methods utilize robot motion parallel to a ground plane and the parallax effect to detect deviations from a flat plane in RGB images. This includes methods which compute homographies between subsequent images~\cite{zhou2006homography, Conrad2010Homography}, and those based on optical flow~\cite{stoffler2000real, camus1996real}. Kumar \textit{et al.}~\cite{kumar2014markov} use an RGB-D camera and robot motion to detect very small obstacles (0.5-2cm) on a planar surface. While such methods work well for moving, wheeled robots, they are not useful for detecting deviations when camera motion is unconstrained or nonexistent, and they assume that the scene is static from frame to frame. Our method operates on a per-frame basis, meaning it does not rely on robot motion nor assume a static scene from frame to frame.

\subsubsection{Stereo-based Methods}
Some methods for obstacle avoidance~\cite{broggi2011stereo, pinggera2016lost, hua2019small} use depth from stereo to determine the distance to the ground plane, and detect an obstacle when it deviates greatly from that ground plane.
Compared to our work, these methods use larger sensors, as they require a stereo baseline, and the algorithms have higher compute requirements.

\subsubsection{Learning-Based Methods}
Many methods for per-frame object detection utilize a large labeled dataset to train a detector via supervised learning~\cite{hua2019small, Liao2022road, ramos2017detecting, gupta2018mergenet}. Other works aim to generally detect any anomaly in an image, relying only on a dataset of anomaly-free images~\cite{di2021pixel, rai2023unmasking}. While many of these methods are effective, the neural networks employed are generally compute intensive, making them unfit for resource constrained applications.

\subsection{Time-Resolved Sensors in Robotics}
Time-resolved time-of-flight sensors which report transient histograms are widely used in robotics to sense the distance to an object. One line of work places these sensors in a distributed manner on robot arms and uses their measurements to ensure safe movement~\cite{escobedo2021contact, Tsuji2019}. There exist methods for calibrating the position of these sensors on a robot arm~\cite{Sifferman2022Geometric}. The sensors' small size makes them well suited for use on small drones for obstacle avoidance~\cite{tsuji2022omnidirectional} and SLAM~\cite{karam2022microdrone}, and for use on very small mobile robots~\cite{niculescu2023nanoslam}. Sifferman \textit{et al.}~\cite{Sifferman2023unlocking} provide a method for recovering planar geometry from the measurements of a time-resolved sensor and build an example robotics application.

\subsection{Low-cost Distance Sensor Transient Histograms}
Transient histograms from low-cost distance sensors have been utilized for human pose estimation~\cite{ruget2022pixels2pose}, object tracking~\cite{Callenberg2021CheapSPAD}, material classification~\cite{becker2023plastic, Callenberg2021CheapSPAD}, depth estimation~\cite{jungerman2022, nishimura2020disambiguating, li2022deltar}, to assist an RGB camera in SLAM~\cite{liu2023multi}, and for multi-view 3D reconstruction~\cite{mu20243d}. While many of these works are similar in spirit to this work, \textit{i.e.}, they aim to take advantage of transient histograms to unlock new capabilities, they do not attempt to solve the same problem of detecting if a surface is planar from a single sensor reading.

\section{Problem Analysis}
\subsection{Background: Transient Histograms}
A time-resolved distance sensor operates by illuminating a wide patch of the scene with a very short pulse of light and capturing the intensity of that light over pico-to-nanosecond timescales as it returns to the sensor after bouncing off of the scene in a \textit{transient histogram} \cite{jungerman2022, gutierrez2022compressive}. Single photon avalanche diodes (SPADs) \cite{niclass2005design, zappa2007principles} are the most mature and widely available technology enabling this type of sensing.

In this work, we utilize the AMS TMF8820 sensor, a SPAD-based distance sensor with $9$ pixels in a $3\times3$ configuration, making it akin to a very low-resolution depth camera. Unlike a typical depth camera, each pixel captures a $128$-bin transient histogram, which can be read out in addition to a per-pixel distance estimate generated by a proprietary algorithm onboard the sensor. We choose this sensor as it reports histograms at a relatively high temporal resolution, with each bin corresponding to $\sim1.2$ cm of depth range. In principle, our method can be applied to any time-resolved optical sensor with a diffuse light source, which currently includes dozens of SPAD-based consumer distance sensor models, in addition to research-grade benchtop setups.

\begin{figure}
    \centering
    \includegraphics{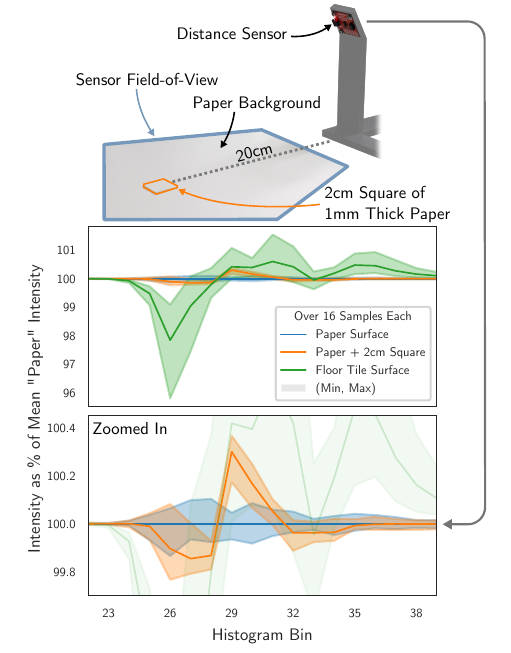}
    \caption{A low-cost distance sensor can distinguish between a flat surface and one with a small piece of heavyweight paper under controlled conditions. The effect of a change in surface photometrics caused by a change to a tile surface is much larger than the effect of a change in geometry caused by the presence of a small piece of paper on a background of the same material.}
    \label{fig:sensitivity}
    \vspace{-1.5em}
\end{figure}

\subsection{Sensor Sensitivity}
Miniature time-resolved sensors can be very sensitive at close distances and have a high signal-to-noise ratio. To demonstrate this with the TMF8820, we perform an experiment in which we place a $20\times20\times1$ mm piece of heavyweight paper on a background of the same material, as shown in Figure~\ref{fig:sensitivity}. Without moving the sensor or background, we capture 16 measurements of the flat surface and 16 measurements of the surface with the small square of paper in the same position 20cm from the sensor. The sensor is set to its default integration period of 230ms, during which it integrates over 4 million laser pulses, giving it an effective frame rate of $\sim$4.3FPS. As shown in Figure~\ref{fig:sensitivity}A, transient histograms between the two scenarios are easily separable when the background surface is highly controlled. This might lead one to expect that the distance sensor would be capable of detecting even sub-mm deviations in a planar surface. However, sensor noise is typically not the limiting factor as discussed in the following subsection.

\subsection{Ambiguity Between Geometry and Albedo}
\label{subsec:ambiguity}

\begin{figure*}
    \centering
    \includegraphics{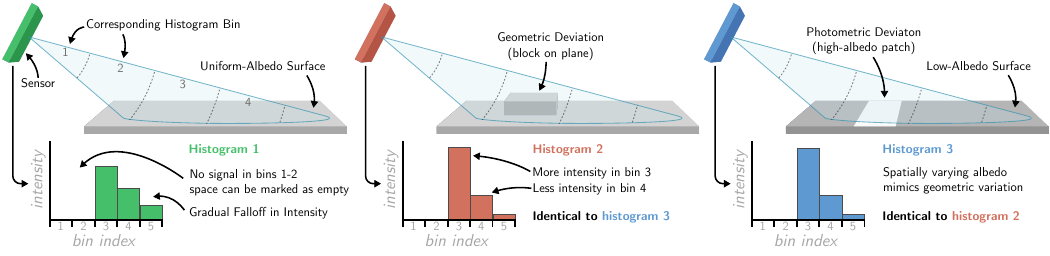}
    \caption{Time-resolved distance sensors exhibit a fundamental ambiguity between geometry and albedo. While a geometric deviation from a flat plane does affect the histogram, a photometric deviation, in the form of a patch with a higher albedo, can affect the histogram in an identical way. This makes the detection of deviation an ill-posed problem aside from geometric variations that violate the space carving assumption. This space carving assumption is weaker at a steep angle of incidence with the plane.}
    \label{fig:ambiguity}
    \vspace{-1.5em}
\end{figure*}

The captured transient histogram is a product of scene geometry, photometric effects (spatially varying scene albedo and reflectance), and sensor intrinsics (laser power, FoV, etc.). As described in previous work~\cite{jungerman2022}, this leads to an ambiguity between scene geometry and photometrics. The intensity of light captured in a given histogram bin is affected by both how \textit{much} surface is within the range corresponding to that bin, and the effective \textit{albedo} of that surface in the wavelength of the light source. From a single measurement, it is impossible to differentiate between a large low-albedo scene patch, and a small high-albedo one.

This ambiguity poses a problem for our goal of detecting deviations in a planar surface. The only constraint that can be placed on the scene from a single measurement is that bins in which no light returns do not contain any geometry, commonly referred to as \textit{space carving}~\cite{kutulakos2000theory, malik2023transient, mu20243d}. This constraint is especially ineffective when the sensor is at a steep angle-of-incidence to the planar surface, \textit{e.g.}, the configuration in Figure~\ref{fig:ambiguity}. Without making assumptions about the plane photometrics or introducing implicit bias into the inference process, it is impossible to detect planar deviations beyond those that violate the space carving constraint. This ambiguity exists regardless of the temporal resolution of the sensor, but is reduced by an increase in spatial resolution (\textit{i.e.}, more pixels, each with a smaller FoV).

We demonstrate that, despite this fundamental ambiguity, it is practical to detect deviations on real-world surfaces by implicitly modeling the expected distribution of surface reflectance, and identifying measurements that are unlikely under the model. We are inspired by works on monocular depth estimation (MDE)~\cite{eigen2014depth, fu2018deep}, which have been successful at predicting ordinal depth despite MDE exhibiting a very similar geometric-photometric ambiguity to our problem. This is possible because real-world data tends to be well-behaved, and adversarial examples that abuse the ambiguity are exceedingly rare. While approaches for MDE take advantage of implicit bias by training a neural network on a large dataset, the relative low-dimensionality of the measurements captured by our sensor means that simpler classical vision techniques can be effective. 

The existence of this ambiguity informs the design of our method. We know that the image formation process cannot be directly inverted to recover geometry, so a model of the expected reflectance properties of the background surface is necessary to make predictions about scene geometry.

\section{Method}
Given a single measurement from a time-resolved distance sensor, we aim to predict whether the geometry of the imaged scene area matches that of a plane viewed from a fixed distance and angle-of-incidence. This is a binary classification problem: given a measurement, classify whether it is an image of a deviation-free plane or not.

At inference time, our method takes as input a set of query histograms $\vb{H} = \{\vb{h}_i = [h_1, h_2, ..., h_b]\}_{i=0}^p$ captured simultaneously by a sensor with $p$ pixels and $b$ bins per histogram. We assume that the histograms have had some form of pile-up correction, like Coates' correction~\cite{coates1968correction} applied, but have not been processed to remove ambient light, which manifests as a constant DC bias in the measured signal~\cite{gupta2019photon}. This is consistent with the transient histograms reported by currently available low-cost distance sensors, including the TMF8820. The output of our method is a likelihood $\ell \in [0, 1]$ of $\vb{H}$ being an image of a deviation-free planar surface viewed from a fixed position, \textit{i.e.}, $p < 0.5$ could mean that there is an object atop the plane, the plane contains a divot downwards, the plane is not being viewed from the fixed distance and angle-of-incidence, or the imaged scene is not a plane at all.

Our method utilizes a dataset $D$ of measurements $\vb{H}$, each of which images a flat planar surface from a fixed distance and angle-of-incidence. This dataset can consist of measurements of a single surface material (to be used \textit{e.g.}, when the surface material is known) or a larger corpus of measurements of multiple surfaces. 
We investigate performance under both dataset types in Section \ref{sec:experimental_results}.
Fitting a model of the deviation-free surface to this dataset consists of two steps: 1) pre-processing $D$ to remove the effect of ambient light and normalize based on albedo, and 2) modeling the distribution of $D$ with a multi-dimensional Gaussian mixture model. This model was chosen based on empirical performance and because it allows lightweight inference and can be trained on only samples of deviation-free planar surfaces.

\subsection{Pre-processing}
We approximate the ambient light level $a_i$ for each histogram $\vb{h}_i$ by calculating kernel density on the values of $\vb{h}_i$ with a Gaussian kernel with bandwidth $\sigma$, and choosing the value with the highest density:
\begin{equation}
    a_i = \underset{x}{\text{argmax}}\sum_{h \in \vb{h_i}} \mathcal{N}(x; h, \sigma)
    \label{eqn:ambient_light_correction}
\end{equation}
\noindent
This serves as a way of estimating the \textit{modal} value in the histogram with more robustness to noise than the mode. For most scenes, the modal bin value is equal to the influence from ambient light, as bins which do not capture light returning from the scene will capture only ambient light, and even a few such bins are typically enough to make it the modal value. In practice, the ideal $\sigma$ varies based on sensor noise, and can be found by searching for the $\sigma$ which minimizes the variation in recovered $a$ over a set of measurements taken under the same ambient light.

Surfaces of different, but uniform albedos will result in histograms of similar shape, but different magnitudes, equivalent to scaling the bin values uniformly. To compensate for this effect, we normalize each histogram to have a unit $L^1$ norm after ambient light has been removed. The pre-processed measurement $\vb{\tilde{H}}$ is given by:
\begin{equation}
    \vb{\tilde{H}} = \left\{ \dfrac{\vb{h}_i - a_i}{\| \vb{h}_i \|_1}\right\}_{i=1}^p
    \label{eqn:pre-processing}
\end{equation}

These pre-processing steps serve to align histograms which have similar shapes, but different magnitudes due to surface albedo, or different DC biases due to ambient light.

\subsection{Gaussian Mixture Model}
\label{subsec:gmm}
We model the distribution of measured histogram values using a multi-dimensional Gaussian mixture model \cite{stauffer1999adaptive}. We flatten each measurement $\vb{\tilde{H}}$ across the pixel dimension to get a one dimensional vector $\vb{\hat{H}} \in \mathbb{R}^k$ where $k=p*b$ containing each $\vb{\tilde{h}}_i \in \vb{\tilde{H}}$ concatenated one after another. Our method does not account for the ordering of the histogram bins or the spatial position of their fields-of-view, so this flattening serves to simplify notation. A Gaussian mixture model is fit to the pre-processed measurements $\vb{\hat{H}}$ in the dataset $D$.

We fit a Gaussian mixture with $c$ components, a per-component per-bin mean vector $\boldsymbol{\mu} \in \mathbb{R}^{c \times k}$, a per-component variance vector $\boldsymbol{\sigma}^2 \in \mathbb{R}^c$, and a per-component weight parameter $\boldsymbol{\alpha} \in \mathbb{R}^c$. This means that a separate Gaussian mixture is fit to each bin, with each Gaussian mixture constrained to use the same variances, weights, and number of components. The likelihood $\ell$ of a given histogram being an image of a flat plane is calculated by taking the joint probability over all bins, with each bin's likelihood calculated by summing all weighted components of the mixture:

\begin{equation}
    \mathcal{L}(\vb{\hat{H}}) = \prod_{i=1}^{k} \sum_{j=1}^{c} \mathcal{N}(\vb{\hat{H}}_i; \boldsymbol{\mu}_{i,j}, \boldsymbol{\sigma}^2_j) * \boldsymbol{\alpha}_j 
    \label{eqn:PDF}
\end{equation}

The Gaussian parameters are estimated over the dataset $D$ of flat planar images using the expectation-maximization algorithm~\cite{dempster1977maximum} to optimize the parameters $\boldsymbol{\mu}, \boldsymbol{\sigma}^2$, and $\boldsymbol{\alpha}$. To choose the number of Gaussian components $c$ to use, we fit models with a range of $c$ values and choose that which minimizes the Akaike information criterion (AIC), given by:

\begin{equation}
    \text{AIC} = 2\rho - 2\ln(\sum_{\vb{\hat{H}} \in D} \mathcal{L}(\vb{\hat{H}}))
    \label{eqn:aic}
\end{equation}

\noindent
where the parameter count $\rho = kc + 2c$. The chosen $c$ varies depending on the contents of $D$.

To predict the likelihood that a query histogram $\vb{H}_{\text{query}}$ is an image of a flat planar surface from the same fixed pose as the measurements in $D$, we perform the same pre-processing steps given in Equation~\ref{eqn:pre-processing} and calculate $\mathcal{L}(\vb{\hat{H}}_{\text{query}})$ as in Equation~\ref{eqn:PDF}. To generate a binary prediction, we apply a threshold to this likelihood, and we label likelihoods above the threshold as ``no deviation'' (the negative class) and those below the threshold as ``deviation'' (the positive class).

\subsection{Baseline Methods}
We compare our method to two baselines, both of which extract from the histograms a single value per-pixel before performing classification on those values. The \textit{proprietary distances} method utilizes the per-pixel distance estimates generated onboard the sensor via a proprietary algorithm. The \textit{histogram peaks} method finds the location of the histogram peak by fitting a cubic curve to the histogram and sampling that curve at a high frequency, following the method of~\cite{Sifferman2023unlocking}. For either method, we plug the distance estimates into the same Gaussian mixture model as described above, but effectively treating each distance estimate as a histogram with one bin, leading to $p$-dimensional feature vector. Empirically, we found this to be the best performing approach for utilizing distance estimates.

\begin{figure}
    \centering
    \includegraphics[]{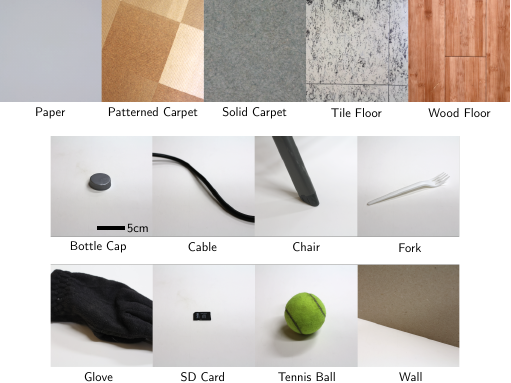}
    \caption{Surfaces (top) and obstacles (bottom) used in obstacle detection experiments.}
    \label{fig:surfaces}
    \vspace{-1.5em}
\end{figure}

\section{Experimental Results}
\label{sec:experimental_results}
We perform real-world experiments with the AMS TMF8820 distance sensor to assess the performance of our method for detecting planar surfaces across many conditions and compare our method to baselines that utilize only a summary statistic for each histogram.

\begin{figure}
    \centering
    \includegraphics[]{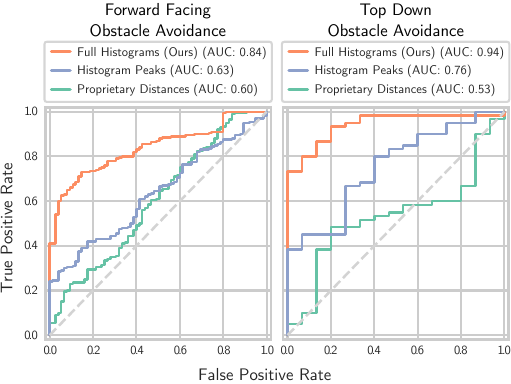}
    \caption{Our method outperforms baseline methods on AUROC on our forward-facing and top-down obstacle detection datasets.}
    \label{fig:ROC_curves}
    \vspace{1.5em}
    \centering
    \includegraphics[]{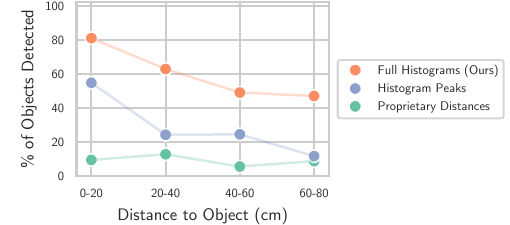}
    \caption{Our method has a higher detection rate than baselines across the entire distance range. Detection threshold is limited to a $<5\%$ false positive rate on the test set for fair comparison.}
    \label{fig:detection_vs_distance}
    \vspace{-1.5em}
\end{figure}

\subsection{Implementation Details}
Measurements are taken with an AMS TMF8820 sensor connected to an Arduino microcontroller via I$^2$C, which forwards the measurements to a connected computer. The sensor is set to 4 million iterations and a measurement period of 230ms, leading to an effective frame rate of $\sim$4.3FPS. KDE kernel bandwidth $\sigma$ in Equation \ref{eqn:ambient_light_correction} is set to 5. Expectation-maximization for Gaussian mixture model fitting is done via the scikit-learn~\cite{scikit-learn} GaussianMixture class. On a mid-range laptop, model fitting takes about 3 seconds on a dataset of 75 measurements, and inference runs at $\sim$108 FPS. We run the sensor in ``low range, high accuracy'' mode, which gives it a maximum range of 120cm. We limit our testing to within 80cm of the sensor and trim all sensor measurements $\vb{h}$ to bins in range $(13, 73)$ after the pre-processing step, corresponding to distances 0-80cm from the sensor. This range makes it more practical to capture measurements of a surface that is planar over the entire sensor FoV.

\subsection{Forward-Facing Obstacle Detection}
\label{subsec:obstacle_detection}

\begin{table*}[]
    \centering
    \begin{tabular}{l c c c c c c c c c}
        & & \multicolumn{8}{c}{Per-Object AUROC $\uparrow$}\\
        \cmidrule(){3-10}
        Method & Overall AUROC $\uparrow$ & Bottle Cap & Cable & Chair & Fork & Glove & SD Card & Tennis Ball & Wall\\ 
        \cmidrule(){1-10}
        Ours (Histograms) & \textbf{0.84} & \textbf{0.83} & \textbf{0.87} & \textbf{0.80} & \textbf{0.84} & \textbf{0.87} & \textbf{0.76} & \textbf{0.86} & \textbf{0.89}\\
        Histogram Peaks & 0.63 & 0.61 & 0.65 & 0.61 & 0.63 & 0.66 & 0.57 & 0.71 & 0.62\\
        Proprietary Distances & 0.60 & 0.58 & 0.55 & 0.59 & 0.54 & 0.63 & 0.70 & 0.52 & 0.68\\
    \end{tabular}
    \caption{Forward-facing obstacle detection results. Our method outperforms baselines overall and across each object.}
    \label{tab:obstacle_detection_results}
    \vspace{-0.5em}
\end{table*}

We capture a large dataset to emulate an obstacle detection scenario for a small mobile robot. We build a mount that holds the sensor 10cm from the ground and at a 60\degree angle-of-incidence to the surface, as in Figure~\ref{fig:sensitivity}, so that the top of its field-of-view is parallel to the ground. We capture measurements of five planar floor surfaces; for each surface, we capture 30 measurements of the surface with no deviations and 10 measurements each of 8 different objects placed at various distances between 10cm and 80cm from the sensor. The surfaces and objects are shown in Figure~\ref{fig:surfaces}. Every capture is under artificial lighting, aside from the wood floor surface, which is under direct sunlight filtered through a window. In total, we capture 550 measurements. The sensor and objects are moved between every capture to ensure diverse coverage of the spatially varying surface. For every measurement, we also capture an RGB and depth image from an Intel RealSense D405 depth-from-stereo camera placed next to the sensor. These depth images are used to manually verify that the objects lie within the sensor's FoV during capture and to label the distance to each object.

\subsubsection{Per-Object Performance}
We use half of the captures of empty surfaces to fit a single surface model ($N=15$ per surface, 75 total) as described in Section~\ref{subsec:gmm}. This emulates a scenario in which the material of the planar surface is constrained, but not exactly known. The model selected by Equation~\ref{eqn:aic} has $c=10$ components. For every query measurement in the test set, Equation~\ref{eqn:PDF} is used to calculate the probability of the sample \textit{not} being a measurement of a deviation-free planar surface. These per-sample probabilities are used to calculate the area under the receiver operating characteristic (AUROC) for the binary classification problem, and those results shown in Table~\ref{tab:obstacle_detection_results}. The ROC curves are shown in Figure~\ref{fig:ROC_curves}. Our method leads to significantly higher AUROC than the baselines, meaning it is better at discriminating between measurements of a flat planar surface and measurements of a surface with an obstacle. When broken down by object, we find that our method is marginally better at detecting large objects (\textit{e.g.}, wall, glove) than small ones (\textit{e.g.}, SD card, bottle cap).

\subsubsection{Performance vs. Distance to Object}
Using the same per-sample probabilities, we utilize the distance labels generated from the depth-from-stereo camera to evaluate obstacle detection performance as a factor of distance to the object. To ensure a fair comparison between methods, we restrict the detection threshold to generate a $<5\%$ false positive rate on the test dataset. Obstacle detection accuracy as a factor of distance is shown in Figure~\ref{fig:detection_vs_distance}. We find that, at all distances, our method outperforms baselines. We also observe that the distance to the object has a larger effect on detection rate than the object size. This may be because the amount of light returning from an object at distance $d$ falls off at a rate of ${1}/{d^2}$, meaning SNR goes down quickly as $d$ increases.

\subsubsection{Performance as a Factor of Training Surfaces}
To evaluate the performance of our method on out-of-distribution surfaces, we perform an experiment in which we vary the surfaces used to fit the surface model and split the AUROC by test surface. This means the model is trained on four surfaces, and must generalize to a fifth. We test three conditions: ``All'' training surfaces, which is the same surface model fit to 75 total measurements of all surfaces, as used for Table~\ref{tab:obstacle_detection_results} and Figures~\ref{fig:ROC_curves} and~\ref{fig:detection_vs_distance}; ``Test only'' training surfaces, in which only 15 measurements from the test surface are used to fit the surface model; and ``All but test'', in which 60 total measurements of all surfaces \textit{except for} the test surface are used to fit the surface model. The results of this experiment are shown in Table~\ref{tab:per_surface}. When the surface model is fit to all five surfaces, performance is uniform across each surface. When fit to and tested on one surface only, our method's performance increases significantly for most surfaces, as per-bin measurements have lower variance and the surface model is able to fit more tightly to the expected distribution. Performance is reduced, however, on the highly spatially varying ``patterned carpet'', as fifteen measurements of the surface alone is not enough to capture a comprehensive range of surface textures. Lastly, when the surface model is fit to all surfaces \textit{but} the test surface, we see a noticeable drop in the performance of our method, with the highest performance decrease present in the highly specular tile floor, and the lowest in the solid carpet. Performance on the tile floor likely suffers because it is specular, while all other surfaces are nearly fully diffuse. Meanwhile, the solid carpet is photometrically more similar to the other four surfaces, leading to better generalization.

\begin{table}
    \centering
    \begin{tabular}{lccc}
        & \multicolumn{3}{c}{AUROC by Training Surfaces $\uparrow$ (\# of samples)}\\
        \cmidrule(){2-4}
        Test Surface & All (75) & Test Only (15) & All but Test (60) \\
        \cmidrule(){1-4}
        Paper & 0.85 & \textbf{0.99} & 0.60 \\
        Patterned Carpet & \textbf{0.83} & 0.65 & 0.70 \\
        Solid Carpet & 0.84 & \textbf{0.92} & 0.79 \\
        Tile Floor & 0.84 & \textbf{0.97} & 0.48 \\
        Wood Floor & 0.83 & \textbf{0.91} & 0.62 \\
        \addlinespace[0.3em]
        (Average) & 0.84 & \textbf{0.89} & 0.71 \\
    \end{tabular}
    \caption{Performance as a factor of the surfaces used to fit the surface model. Performance is highest when the test surface is the only surface in the training set (Test Only), is lower when the test surface is one of five in the training set (All), and is much lower when the test surface is not in the training set (All but Test).}
    \label{tab:per_surface}
    \vspace{-1.5em}
\end{table}

\subsection{Top-Down Obstacle Detection}
We also apply our method to a top-down obstacle detection scenario, as might be utilized by \textit{e.g.}, a drone finding a safe spot to land, or a manipulator finding a safe spot to place an object. In this experiment, the sensor is placed 28cm above the ground facing straight downwards. We use the ``solid carpet'' material from Section~\ref{subsec:obstacle_detection} and the bottle cap, cable, fork, glove, SD card, and tennis ball objects. Similar to the forward-facing obstacle detection experiment, 30 total measurements are taken of the empty surface, half of which are reserved for fitting the surface model. Ten pictures of each object are taken, with the object moved within the FoV and the sensor moved relative to the surface for each capture. The model selected by Equation~\ref{eqn:aic} has $c=4$ components. The resulting ROC curves of each method are shown in Figure~\ref{fig:ROC_curves}. Our results echo those of the forward-facing obstacle detection experiment: utilizing the entirety of the histograms makes it much easier to discern between planar surfaces and non-planar surfaces. Performance in this setting at this distance is similar to that achieved in our forward-facing obstacle detection experiment, when only the solid carpet surface is used to fit the surface model.

\subsection{Cliff Detection}
In principle, our method can detect any deviation in a planar surface, \textit{e.g.}, a protrusion upwards, cliff, or ledge. To evaluate cliff detection performance, we gather a dataset in which the sensor is placed atop a wooden table, using the same forward-facing mount. We use 15 measurements of the empty tabletop to fit a model for the surface, and the sensor is placed at varying distances from the edge of the table in the range (5cm, 75cm) in 5cm increments. At each distance, four measurements are taken from various positions. An additional 15 measurements of the empty planar surface are reserved for testing to test false positive rate. Our method is able to detect cliffs 100\% of the time up to 35cm away with no false positives. Cliffs 45cm away or further are not detected. Under the same conditions, the histogram peak-based method detects cliffs up to 25cm, while the proprietary distance-based method is only effective up to 10cm.


\subsection{Ablation Study}
We perform an ablation study to assess the importance of each aspect of our method on the forward-facing obstacle detection dataset. The results of this study are shown in Table~\ref{tab:ablation}. We find that each aspect of our method has some effect on performance, with exclusion of both pre-processing steps leading to a drop of 0.8 AUROC, and limiting the Gaussian mixture model to one component leading to a large drop of 0.21 AUROC. We believe that this is because bin values tend to be multi-modal, especially when the surface model is fit to a multiple-surface dataset, as different surfaces' varying reflectance properties mean that the shape of the histogram is consistent within a surface (aside from patterned surfaces), but varies between surfaces. By limiting the model to a single Gaussian component, the varying histogram shapes are not effectively modeled.

\begin{figure}
    \centering
    \includegraphics[]{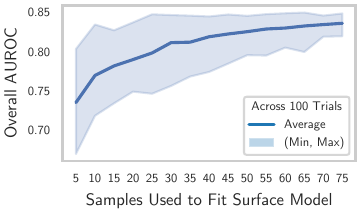}
    \caption{Effect of varying the number of samples used to fit the surface model on AUROC on our forward-facing obstacle detection dataset. The shaded region represents the minimum and maximum over 100 samples, while the solid line represents the average.}
    \label{fig:vary_num_samples}
    \vspace{-0.5em}
\end{figure}

We perform an additional study in which we vary the number of samples used to fit the surface model for forward-facing obstacle detection. For a given number of samples, we pull an even number of samples from each of the five surfaces. For each number of samples, we repeat the experiment 100 times, each with a randomly sampled dataset for surface model fitting and the same test set. The results of this experiment are shown in Figure~\ref{fig:vary_num_samples}. We find that performance begins to level out as we approach 15 samples per surface (75 total). Reasonable performance is still possible with fewer samples per surface, \textit{e.g.}, 5 samples per surface (25 total) yields an average AUROC of 0.80.

\begin{table}
    \centering
    \begin{tabular}{lc}
        Method & AUROC $\uparrow$\\
        \cmidrule(){1-2}
        Base & 0.84 \\
        \addlinespace[0.3em]
        No Ambient Light Correction ($a_i = 0$) & 0.82 \\
        No Normalization (modify Eqn. \ref{eqn:pre-processing}) & 0.78 \\
        No Norm. or Ambient Light Correction (skip Eqn. \ref{eqn:pre-processing}) & 0.76 \\
        Limit to one Gaussian Component ($c=1$) & 0.63 \\
        No Norm, No ALC, \& Limit to One Component & 0.53
    \end{tabular}
    \caption{Ablation study results on obstacle detection dataset.}
    \label{tab:ablation}
    \vspace{-1.0em}
\end{table}

\section{Example Application}
We build an example application for our method in which a mobile robot is equipped with three distance sensors in a forward-facing configuration, providing a wide field-of-view. The sensor is in the same position relative to the ground as in our forward-facing obstacle detection and cliff detection experiments. We assume that the robot begins in an obstacle-free area and can safely drive forward for 10 seconds to characterize the surface. After capturing 30 measurements per sensor to characterize the ground surface and fit a surface model, the robot is able to avoid obstacles and cliffs using measurements from the sensors alone. The maximum range at which obstacles are detected is configurable by trimming the histogram bins per sensor during inference. A visualization of the application is shown in Figure~\ref{fig:demo}. See the supplementary material for a video demonstration.

\begin{figure}
    \centering
    \includegraphics[]{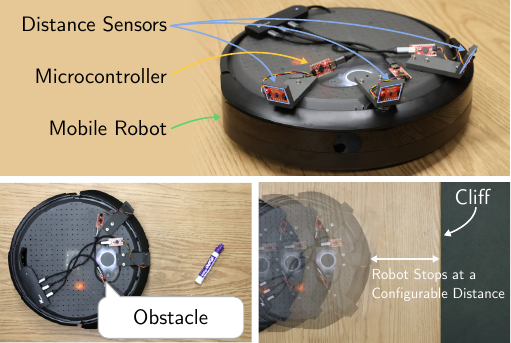}
    \caption{Example application of our method being applied to mobile robot obstacle avoidance. The robot is equipped with three distance sensors. After characterizing the surface, it is able to avoid obstacles and cliffs in its path with a configurable buffer distance.}
    \label{fig:demo}
    \vspace{-1.5em}
\end{figure}

\section{Limitations and Conclusion}
While we have shown that real-world deviations can be detected with reasonable accuracy, our method is still subject to the photometric-geometric ambiguity described in Section~\ref{subsec:ambiguity}. Additionally, because of the presence of photometric effects, our method is most effective when the surface is well-known, and it is much less effective when an observed surface has not been previously seen. Future work should investigate ways to overcome this limitation by, \textit{e.g.}, creating a general-purpose surface model by fitting to a large dataset of surfaces or utilizing an explicit photometric model to set bounds on the expected photometric deviations of the surface. Lastly, we assume that the relative orientation and distance to the planar surface is fixed. Future work should investigate the possibility of detecting surface geometry regardless of relative sensor pose and investigate robustness to subtle changes in sensor pose due to robot motion.

This work provides a way to extend the capabilities of distance sensors with no additional hardware and minimal compute overhead. We look forward to future robotics applications that make use of our method to improve robot sensing, particularly in resource constrained scenarios. 


\bibliography{references}
\bibliographystyle{IEEEtran}

\end{document}